\documentclass[12pt]{article}

\usepackage{amsmath}
\usepackage{geometry}
\usepackage{setspace}
\usepackage{parskip}
\usepackage{graphicx}
\usepackage{authblk}
\usepackage{algorithm}
\usepackage{enumitem}
\usepackage{bm}
\usepackage{algpseudocode}
\usepackage{subcaption}
\usepackage{amsfonts}
\usepackage[
  backend=biber,
  style=numeric
]{biblatex}
\usepackage[colorlinks=false, pdfborder={0 0 0}]{hyperref}

\title{\textbf{PDE-aware Optimizer for Physics-informed Neural Networks}}

\author{\textbf{Hardik Shukla, Manurag Khullar, Vismay Churiwala}\\
ENM 5320: AI4Science}
\date{}

\begin{document}

\maketitle

\begin{abstract}
Physics-Informed Neural Networks (PINNs) have emerged as a powerful framework for solving partial differential equations (PDEs) by embedding physical constraints into the loss function. However, standard optimizers such as Adam often struggle to balance competing loss terms, particularly in stiff or ill-conditioned systems. In this work, we propose a PDE-aware optimizer that adapts parameter updates based on the variance of per-sample PDE residual gradients. This method addresses gradient misalignment without incurring the heavy computational costs of second-order optimizers such as SOAP. We benchmark the PDE-aware optimizer against Adam and SOAP on 1D Burgers', Allen–Cahn and Korteweg–de Vries(KdV) equations. Across both PDEs, the PDE-aware optimizer achieves smoother convergence and lower absolute errors, particularly in regions with sharp gradients. Our results demonstrate the effectiveness of PDE residual-aware adaptivity in enhancing stability in PINNs training. While promising, further scaling on larger architectures and hardware accelerators remains an important direction for future research.
\end{abstract}

\section{Introduction}

A significant advancement in scientific machine learning is physics-informed machine learning, which integrates physical laws and domain-specific constraints directly into the learning process. This integration can be achieved in various ways, including modifications to key components such as the model architecture, loss functions, optimisation algorithms, and hyperparameter tuning \cite{Karniadakis2021PhysicsInformedML, Raissi2019PINNs}. A particularly effective strategy within physics-informed machine learning is embedding physical principles such as conservation laws, symmetries, invariances, and equivariances into the machine learning framework. Among the most flexible and widely adopted approaches is the use of tailored loss functions. These loss functions encode the physical constraints as soft penalties, guiding the model to learn solutions that are not only data-driven but also consistent with known physical laws. This approach has led to the development of Physics-Informed Neural Networks (PINNs), which have been successfully applied to both forward and inverse problems involving partial differential equations (PDE).

While PINNs offer conceptual simplicity and implementation flexibility, training them effectively for stiff or ill-conditioned PDEs remains challenging. Traditional first-order optimization methods, such as Adam, often struggle with convergence or require extensive tuning. To address these issues, researchers have turned to second-order optimization techniques, which utilize curvature information to make more informed updates to the model parameters \cite{Wang2025GradientAlignment, Wang2021GradientPathologies}. Second-order optimizers such as Second-Order Adaptive Optimisation (SOAP) and Second-order Clipping Shampoo (SHAMPOO) have demonstrated superior convergence properties in training PINNs, particularly on complex PDE problems \cite{Wang2025GradientAlignment}. These methods leverage second-order information, such as approximations of the Hessian or the Fisher information matrix, to scale the gradient updates more precisely along each parameter direction, resulting in improved training stability and accuracy. However, these benefits come at a significant cost.

The primary drawback of second-order methods lies in their high computational and memory overhead. Maintaining and updating curvature matrices—often of size proportional to the number of parameters squared—is resource-intensive, especially for deep networks or large-scale problems \cite{Martens2010HessianFree,Martens2015KFAC}. Moreover, operations such as matrix inversion or eigendecomposition add to the computational complexity \cite{RoostaKhorasani2016SubsampledNewton}. Despite these challenges, ongoing research aims to make second-order methods more tractable through matrix sketching, low-rank, and block-diagonal approximations \cite{Gupta2018Shampoo,Yao2021AdaHessian}.

In this project, we aim to develop a PDE-aware optimizer that can:
\begin{itemize}
    \item Align gradients from multiple loss terms (PDE residual, boundary, and initial conditions),
    \item Adaptively scale updates to prevent unstable PDE gradients from dominating, and
    \item Avoid expensive operations like Hessian inversion, making it practical for real-world problems.
\end{itemize}

\section{Loss terms in PINNs}

We consider a general formulation of a parabolic partial differential equation (PDE) given by
\begin{equation}
\mathbf{u}_t + \mathcal{D}[\mathbf{u}] = \mathbf{f},
\qquad (t,\mathbf{x}) \in [0,T]\times\Omega \subset \mathbb{R}^{1+d}.
\label{eq:pde}
\end{equation} 
where $\Omega$ is a bounded subset of $\mathbb{R}^d$ with a sufficiently regular boundary denoted by $\partial \Omega$. Here, $\mathcal{D}[\cdot]$ represents a linear or nonlinear differential operator, and $\mathbf{u}(t, \mathbf{x})$ denotes the unknown solution to be approximated.

The PDE is supplemented with the following initial and boundary conditions:
\begin{subequations}\label{eq:ic_bc}
\begin{align}
\mathbf{u}(0,\mathbf{x}) &= \mathbf{g}(\mathbf{x}), & \mathbf{x} &\in \Omega,\\
\mathcal{B}[\mathbf{u}](t,\mathbf{x}) &= 0,           & t &\in [0,T],\;
                                            \mathbf{x} \in \partial\Omega.
\end{align}
\end{subequations}

Here, $\mathbf{f}$ and $\mathbf{g}$ are given functions with appropriate smoothness, and $\mathcal{B}[\cdot]$ is a boundary operator which may represent Dirichlet, Neumann, Robin, or periodic boundary conditions. To approximate the solution $\mathbf{u}(t, \mathbf{x})$, we use a deep neural network $u_\theta(t, \mathbf{x})$ parameterized by $\theta$, which includes all trainable parameters such as weights and biases. When a smooth activation function is employed, the neural network $u_\theta$ yields a differentiable approximation that can be evaluated at any point $(t, \mathbf{x})$. Moreover, automatic differentiation enables efficient computation of the necessary derivatives with respect to both the input variables and network parameters.

Using this framework, we define the residuals for the PDE, initial condition, and boundary condition as follows:

\begin{equation}
\mathcal{R}_{\text{int}}[u_\theta](t, \mathbf{x}) = \frac{\partial u_\theta}{\partial t}(t, \mathbf{x}) + \mathcal{D}[u_\theta](t, \mathbf{x}) - f(\mathbf{x}), \quad (t, \mathbf{x}) \in [0, T] \times \Omega,
\end{equation}

\begin{equation}
\mathcal{R}_{\text{bc}}[u_\theta](t, \mathbf{x}) = \mathcal{B}[u_\theta](t, \mathbf{x}), \quad (t, \mathbf{x}) \in [0, T] \times \partial \Omega,
\end{equation}

\begin{equation}
\mathcal{R}_{\text{ic}}[u_\theta](\mathbf{x}) = u_\theta(0, \mathbf{x}) - g(\mathbf{x}), \quad \mathbf{x} \in \Omega.
\end{equation}

The physics-informed neural network (PINN) is then trained by minimising the following composite empirical loss function, which aggregates the residuals from the initial condition, boundary condition, and PDE:

\begin{equation}
\mathcal{L}(\theta) = 
\underbrace{\frac{1}{N_{\text{ic}}} \sum_{i=1}^{N_{\text{ic}}} \left| \mathcal{R}_{\text{ic}}[u_\theta](\mathbf{x}_{\text{ic}}^i) \right|^2}_{\mathcal{L}_{\text{ic}}(\theta)} + 
\underbrace{\frac{1}{N_{\text{bc}}} \sum_{i=1}^{N_{\text{bc}}} \left| \mathcal{R}_{\text{bc}}[u_\theta](t_{\text{bc}}^i, \mathbf{x}_{\text{bc}}^i) \right|^2}_{\mathcal{L}_{\text{bc}}(\theta)} + 
\underbrace{\frac{1}{N_r} \sum_{i=1}^{N_r} \left| \mathcal{R}_{\text{int}}[u_\theta](t_r^i, \mathbf{x}_r^i) \right|^2}_{\mathcal{L}_r(\theta)}.
\end{equation}

\section{Gradient Misalignment in PINNs}

With multiple loss terms, a critical challenge in training PINNs is the conflict among gradients arising from them. This gradient misalignment significantly hampers training efficiency and convergence. The issue is especially severe in multi-physics problems or stiff PDE systems where multiple constraints such as the PDE residual, boundary conditions, and initial conditions must be satisfied simultaneously.

The two distinct types of gradient conflicts \cite{Wang2025GradientAlignment}:

\begin{itemize}
    \item \textbf{Type I:} This occurs when gradients from different loss terms are directionally aligned but have vastly different magnitudes. In this case, smaller gradients may be overshadowed by dominant ones, leading the optimizer to focus excessively on a subset of constraints while neglecting others.
    
    \item \textbf{Type II:} Here, the gradients have comparable magnitudes but point in opposing directions. This destructive interference prevents meaningful progress toward satisfying all physical constraints and results in erratic optimisation trajectories.
\end{itemize}

Both types of conflicts degrade training performance. Type I leads to loss term dominance, where only a few constraints are enforced, while Type II causes oscillations or divergence in parameter updates, particularly during the early stages of training.

Traditional gradient-based optimisation methods, such as Adam, typically assume a unified descent direction. However, in PINNs, different physics-based loss components may push the parameter updates in conflicting directions. Without accounting for these misalignments, the optimizer will likely satisfy some constraints while ignoring others, suffer from poor convergence rates, and fail to find meaningful solutions.

\begin{figure} [H]
    \centering
    \includegraphics[width=0.5\linewidth]{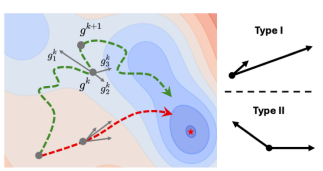}
    \caption{Type I: The irregular green trajectory illustrates how the optimisation struggles when facing two types of gradient conflicts. Type II: The red trajectory shows how appropriate preconditioning through Second-order information could mitigate these conflicts by aligning gradients both within and between optimisation steps.}
    \label{fig:enter-label}
\end{figure}

\section{PDE-Aware Optimizer}
To resolve the gradient misalignment issues, we introduce a variant of Adam whose
second–moment accumulator is driven by per-sample physics-residual gradients. We compute the variance of the per-sample PDE gradients and divide each momentum component by the square root of that variance. Weights whose residual gradients fluctuate strongly across collocation points, therefore, take smaller steps, while those with consistent gradients move farther, giving the optimizer a built-in sensitivity to stiff regions, sharp fronts, or other difficult parts of the solution. Because the first-moment term is the batch-average of those same residual gradients, the update direction is aligned with the dominant physics signal and is less disturbed by initial- or boundary-condition terms that may point elsewhere.


For a mini-batch of $B$ collocation points
$\{x_i\}_{i=1}^{B}$ we compute
\begin{equation}
\bm{g}_i \;=\;
\nabla_{\bm{w}}\,R_{\mathrm{pde}}(x_i;\bm{w}_t),
\qquad
\overline{\bm{g}}
\;=\;
\frac{1}{B}\sum_{i=1}^{B}\bm{g}_i .
\end{equation}

\begin{align}
\bm{m}_t &=
  \beta_1\,\bm{m}_{t-1} + (1-\beta_1)\,\overline{\bm{g}},
\\
\bm{v}_t &=
  \beta_2\,\bm{v}_{t-1}
  +\frac{1-\beta_2}{B}\sum_{i=1}^{B}\bm{g}_i^{\odot 2},
\label{eq:variance-update}
\\
\bm{w}_{t+1} &=
  \bm{w}_t
  -\eta\,\frac{\bm{m}_t}{\sqrt{\bm{v}_t}+\epsilon},
\label{eq:update}
\end{align}
where $\odot$ denotes the element-wise square.
Because each coordinate $m_t^{(j)}$ is scaled by
$\bigl(\sqrt{v_t^{(j)}}+\epsilon\bigr)^{-1}$, the step size is
\emph{smaller} where the residual gradient varies strongly across the
batch and \emph{larger} where it is coherent, focusing adaptation on
difficult physics regions rather than on the total loss.

\begin{algorithm}[H]
\caption{PDE-Aware Optimizer for training PINNs}
\label{alg:pao}
\begin{algorithmic}[1]
  \Require initial parameters $\bm{w}_0$,
          learning rate $\eta$,
          decay rates $\beta_1,\beta_2$,
          batch size $B$,
          small constant $\epsilon$
  \Ensure  trained parameters $\bm{w}_T$
  \State $\bm{m}_{-1}\gets\bm{0}$ \Comment{first moment}
  \State $\bm{v}_{-1}\gets\bm{0}$ \Comment{second moment}
  \For{$t = 0$ \textbf{to} $T-1$}
    \ForAll{$x_i$ in mini-batch}
       \State $\bm{g}_i\gets
              \nabla_{\bm{w}}\,R_{\mathrm{pde}}(x_i;\bm{w}_t)$
    \EndFor
    \State $\displaystyle
           \overline{\bm{g}}\gets
           \frac1B\sum_{i=1}^{B}\bm{g}_i$
    \State $\bm{m}_t\gets
           \beta_1\bm{m}_{t-1} + (1-\beta_1)\,\overline{\bm{g}}$
    \State $\displaystyle
           \bm{v}_t\gets
           \beta_2\bm{v}_{t-1}
           +\frac{1-\beta_2}{B}\sum_{i=1}^{B}\bm{g}_i^{\odot 2}$
    \State $\tilde{\bm{m}}_t\gets
           \bm{m}_t \,/\,\bigl(\sqrt{\bm{v}_t}+\epsilon\bigr)$
    \State $\bm{w}_{t+1}\gets\bm{w}_t-\eta\,\tilde{\bm{m}}_t$
  \EndFor
\end{algorithmic}
\end{algorithm}
\begin{figure}[t]
  \centering
  \includegraphics[width=.55\linewidth]{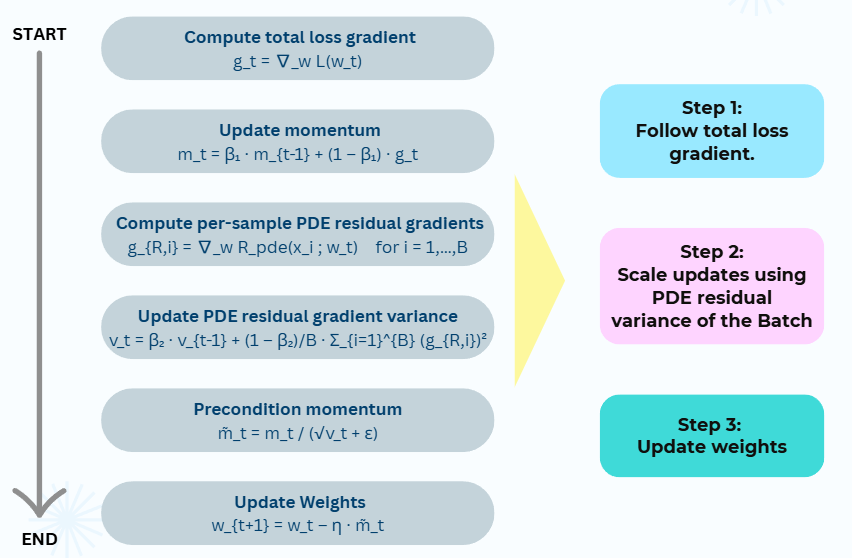}
  \caption[PAO flow chart]{Flow chart of the \textbf{PDE-Aware Optimizer}.
  The first moment $\bm{m}_t$ is built from batch-averaged PDE‐residual
  gradients; the second moment $\bm{v}_t$ tracks their element-wise
  variance, so the pre-conditioned step
  $\bm{w}_{t+1}=\bm{w}_t-\eta\,\bm{m}_t/(\sqrt{\bm{v}_t}+\epsilon)$
  automatically shrinks learning rates in stiff regions and enlarges them
  where the residual is smooth.}
  \label{fig:pao-flow}
\end{figure}

\section{Experimental Setup}

Following common practice in PINNs literature, we benchmark the proposed
Physics-Aware Optimizer on three canonical 1-D PDEs:
Burgers’ convection–diffusion equation, the Allen–Cahn
reaction–diffusion equation and the Korteweg–De Vries wave equation.  
\paragraph{PDE benchmarks:}\textbf{Burgers}
 , \textbf{Allen–Cahn} and \textbf{Korteweg–De Vries} on the common domain
$(x,t)\in[-1,1]\times[0,1]$.
The interior residual minimised by the PINN is
\[
\mathcal{R}(x,t;\bm{w}) =
\begin{cases}
\partial_t u_\theta + u_\theta\,\partial_x u_\theta - \nu\,\partial_{xx} u_\theta,
  &\text{Burgers},\\[6pt]
\partial_t u_\theta - \epsilon\,\partial_{xx} u_\theta - f(u_\theta),
  &\text{Allen--Cahn},\\[6pt]
\partial_t u_\theta + u_\theta\,\partial_x u_\theta + \mu\,\partial_{xxx} u_\theta,
  &\text{Korteweg--de Vries (KdV)}.
\end{cases}
\]

\paragraph{Optimizer benchmark:} Standard Adam, the second-order SOAP, and PDE-aware Optimizer
\paragraph{Model:} A single MLP: input $(x,t)\in\mathbb{R}^2$; hidden $3\times64$ \texttt{tanh} layers; output $u_\theta(x,t)\in\mathbb{R}$

\paragraph{Collocation sampling.}
We generate training data as follows:
\begin{itemize}[leftmargin=*]
  \item \textbf{Interior (PDE residual):} $N_{\text{int}} = 10{,}000$ collocation points sampled uniformly over the spatio-temporal domain $(x, t) \in [x_{\min}, x_{\max}] \times [t_{\min}, t_{\max}]$.
  \item \textbf{Initial condition:} $N_{\text{ic}} = 1{,}000$ points sampled along $t = t_{\min}$ with $x \in [x_{\min}, x_{\max}]$ evenly spaced.
  \item \textbf{Boundary condition:} $N_{\text{bc}} = 1{,}000$ points sampled with $t \in [t_{\min}, t_{\max}]$, and $x = x_{\min}$ or $x = x_{\max}$ fixed on either boundary.
\end{itemize}

\paragraph{Training:}
\textbf{Batch size} $B = 1{,}024$; \textbf{epochs} $T = 10{,}000$.

\begin{figure}[H]
    \centering
    \includegraphics[width=0.6\linewidth]{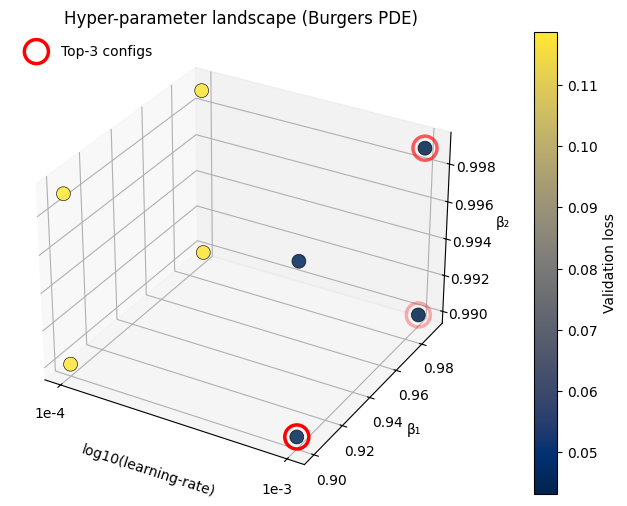}
    \caption{Validation-loss landscape over the $(\eta,\beta_{1},\beta_{2})$ grid for Burgers PDE.}
    \label{fig:hyper-tuning}
\end{figure}

\paragraph{Evaluation Metric:}
Relative $L^{2}$ error on a $400\times400$ grid:
\[
\text{Rel-}L^{2} =
\Bigl(
  \tfrac{\sum (u_\theta-u_{\text{exact}})^2}
        {\sum u_{\text{exact}}^2}
\Bigr)^{1/2}.
\]
\paragraph{Hyper-parameter tuning:}
We carried out a full $2\times2\times2$ grid search over the learning-rate $\eta\!\in\!\{10^{-3},10^{-4}\}$ and moments $\beta_1\!\in\!\{0.9,0.99\}$, $\beta_2\!\in\!\{0.99,0.999\}$ for our PDE-aware optimizer on the 1-D Burgers equation. We chose to begin at the default values of the Adam optimizer and vary the hyperparameters from there. The rationale behind the choice for $\beta_2$ was to increase $1-\beta_2$ since that term would determine how much the corresponding residual term would be weighted in $v_t$.
Figure \ref{fig:hyper-tuning} visualizes the validation loss surface for the eight configurations.  
The best setting—$\eta=10^{-3},\;\beta_1=0.99,\;\beta_2=0.99$—achieved an average validation loss of $4.32\times10^{-2}$, with the two closest neighbors ($\eta=10^{-3},\beta_1=0.99,\beta_2=0.999$ and $\eta=10^{-3},\beta_1=0.9,\beta_2=0.99$) trailing by less than $3\times10^{-4}$.

\section{Results}
\subsection {1D Burgers PDE}
\begin{figure}[H]
    \centering

    \begin{subfigure}[t]{0.24\textwidth}
        \centering
        \includegraphics[width=\linewidth]{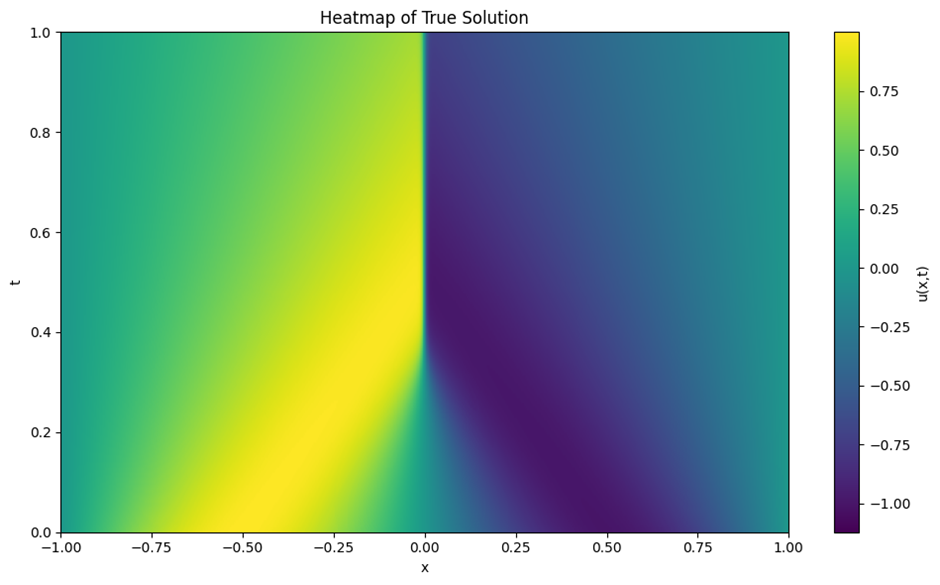}
        \caption{True}
    \end{subfigure}
    \hfill
    \begin{subfigure}[t]{0.24\textwidth}
        \centering
        \includegraphics[width=\linewidth]{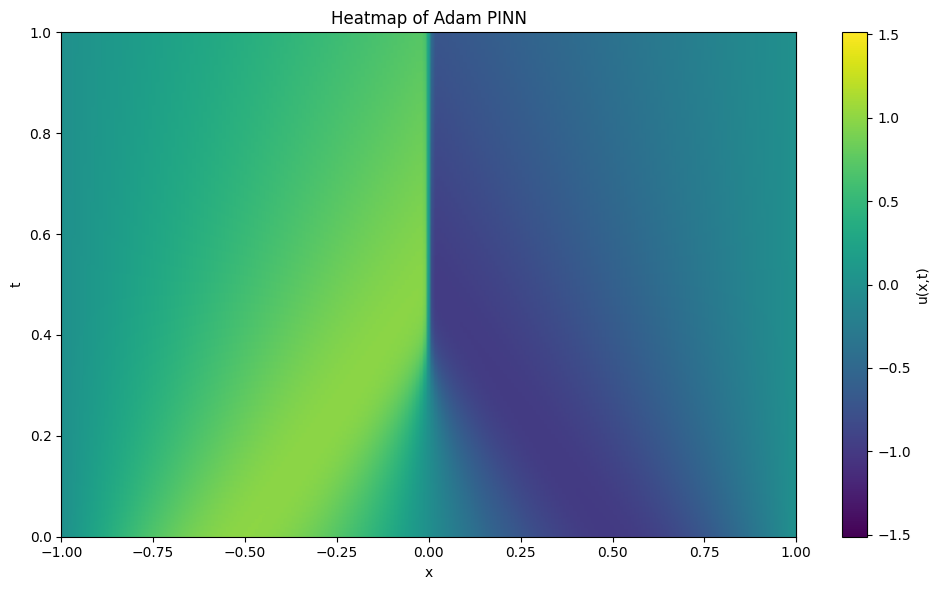}
        \caption{Adam}
    \end{subfigure}
    \hfill
    \begin{subfigure}[t]{0.24\textwidth}
        \centering
        \includegraphics[width=\linewidth]{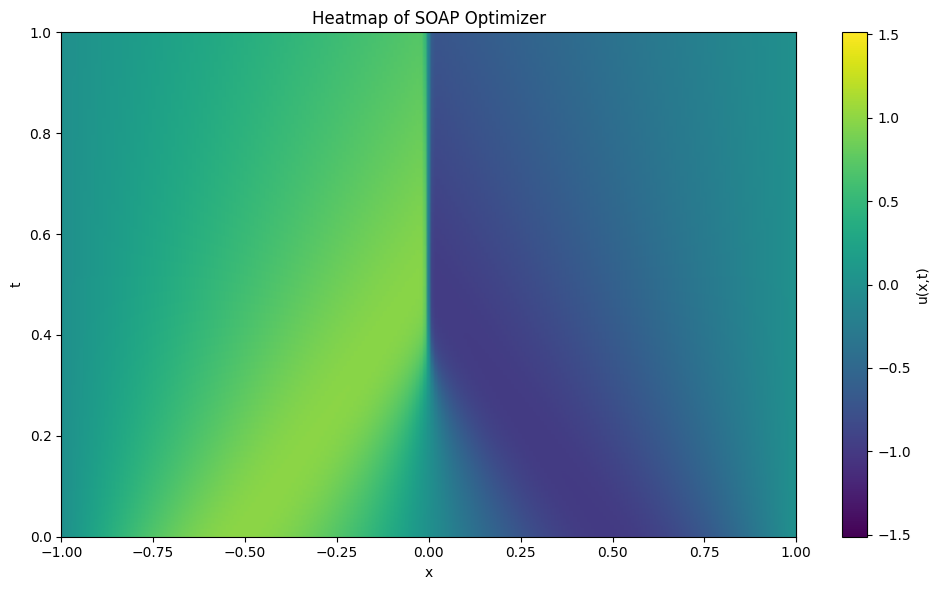}
        \caption{SOAP}
    \end{subfigure}
    \hfill
    \begin{subfigure}[t]{0.24\textwidth}
        \centering
        \includegraphics[width=\linewidth]{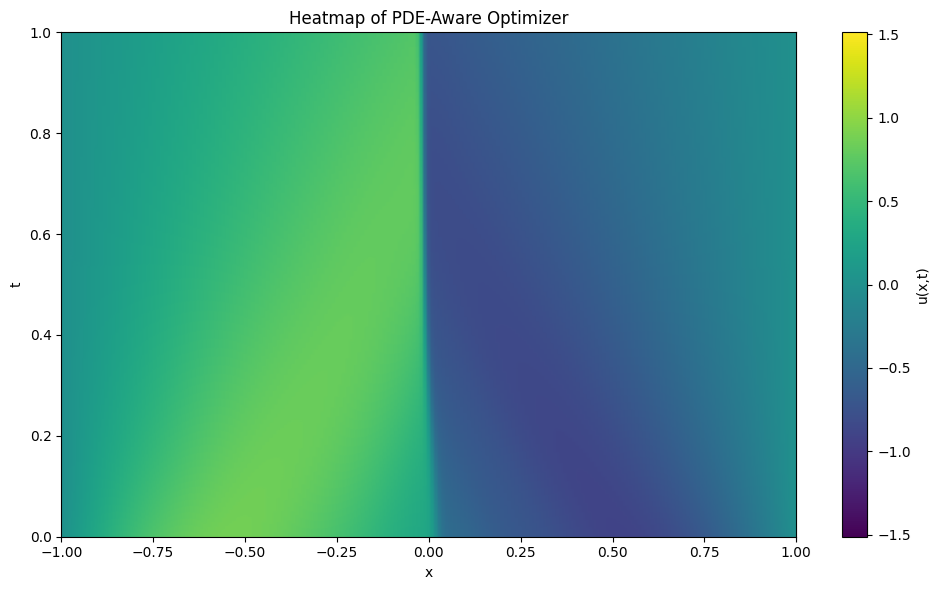}
        \caption{PDE-aware}
    \end{subfigure}

    \label{fig:burger-heatmap-comparison}
   \caption{Heatmap comparison of Adam, SOAP, and PDE-aware optimizers on the Burgers' equation}
   \begin{figure}[H]
    \centering

    \begin{subfigure}[t]{0.32\textwidth}
        \centering
        \includegraphics[width=\linewidth]{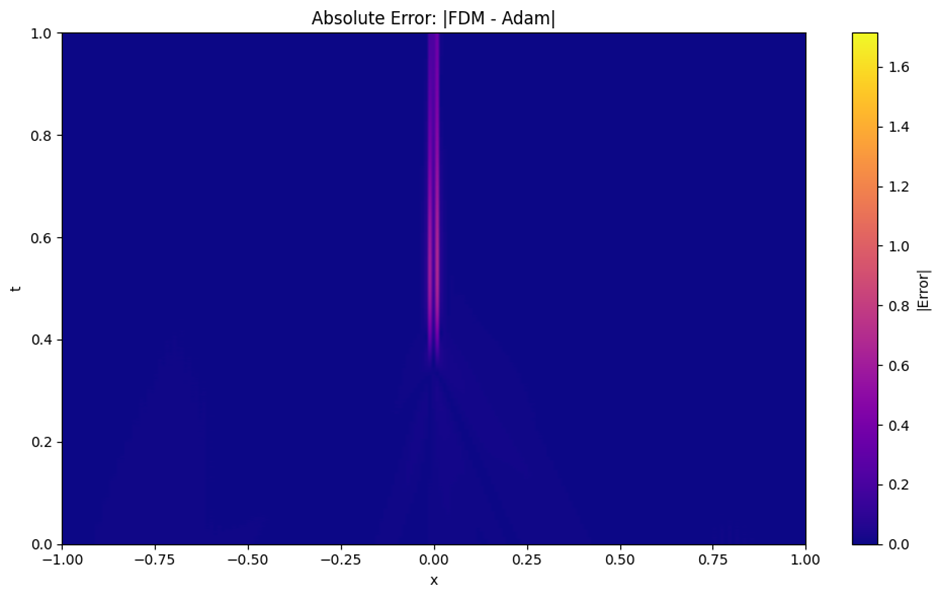}
        \caption{Adam}
    \end{subfigure}
    \hfill
    \begin{subfigure}[t]{0.32\textwidth}
        \centering
        \includegraphics[width=\linewidth]{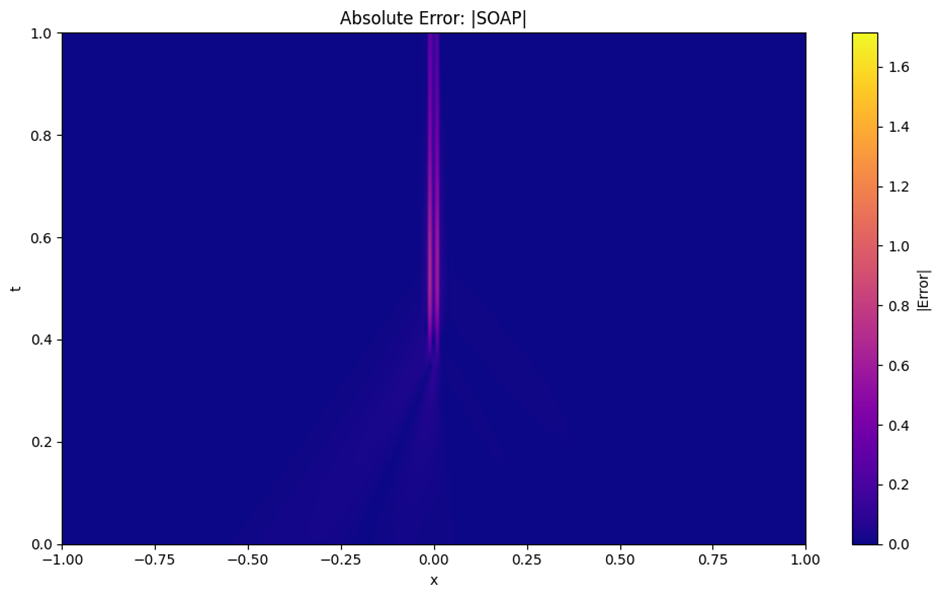}
        \caption{SOAP}
    \end{subfigure}
    \hfill
    \begin{subfigure}[t]{0.32\textwidth}
        \centering
        \includegraphics[width=\linewidth]{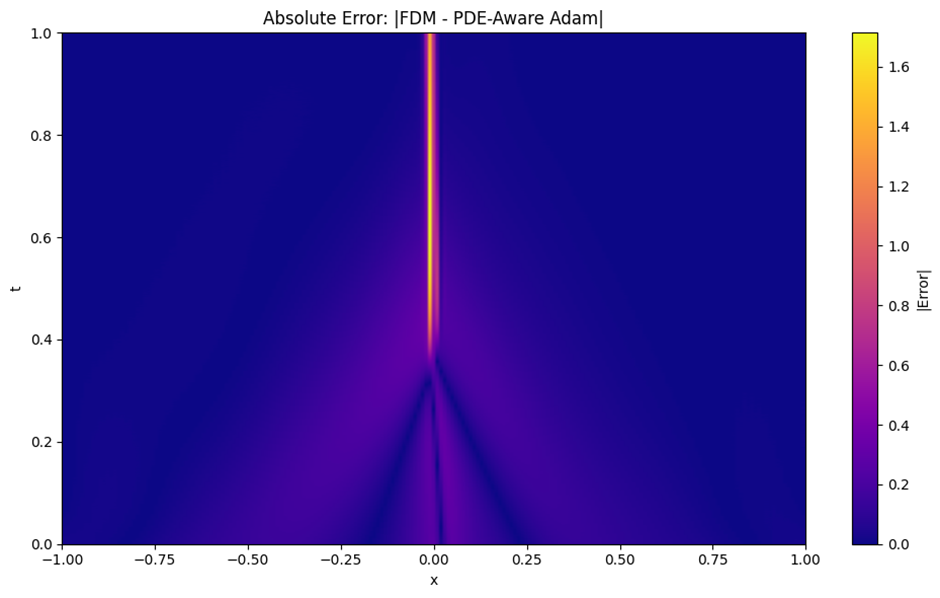}
        \caption{PDE-aware}
    \end{subfigure}

    \label{fig:burger-absolute-error}
       \caption{Absolute error ($|u_{\text{PINN}} - u_{\text{FDM}}|$) for Burgers’ equation across different optimizers}
\end{figure}
\end{figure}
\begin{figure}[H]
    \centering
    \includegraphics[width=0.35\linewidth]{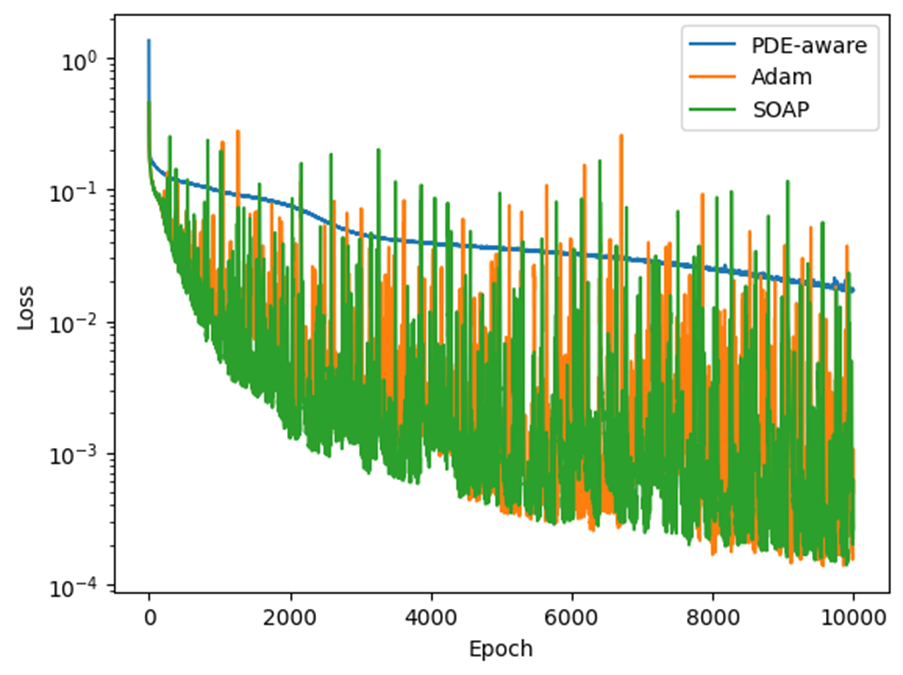}
    \caption{Training loss over epochs for Burgers’ equation. PDE-aware optimizer shows smooth convergence, while Adam and SOAP exhibit faster but more oscillatory behavior.}
    \label{fig:burgers-loss}
\end{figure}

\subsection{1D Allen–Cahn PDE}

\begin{figure}[H]
    \centering
    \begin{subfigure}[t]{0.24\textwidth}
        \centering
        \includegraphics[width=\linewidth]{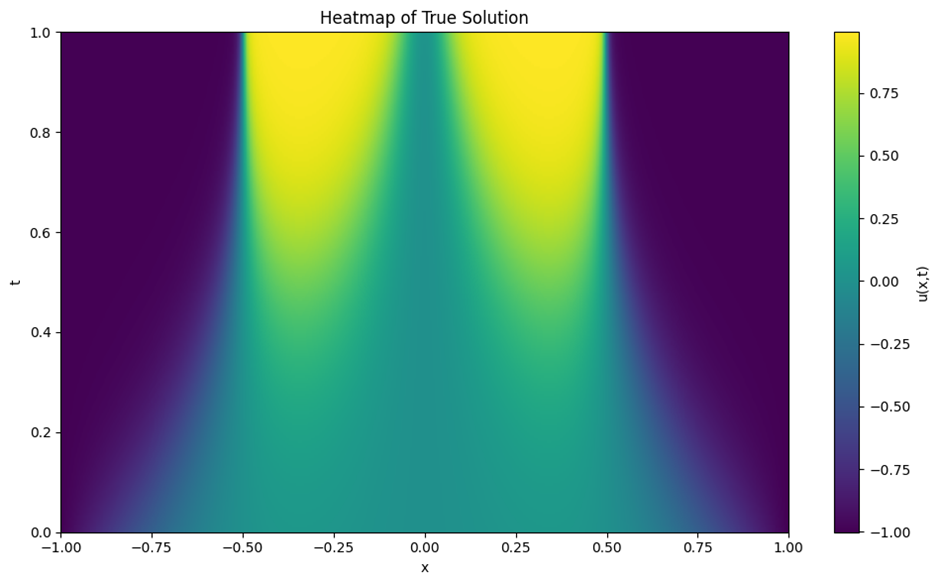}
        \caption{True}
    \end{subfigure}
    \hfill
    \begin{subfigure}[t]{0.24\textwidth}
        \centering
        \includegraphics[width=\linewidth]{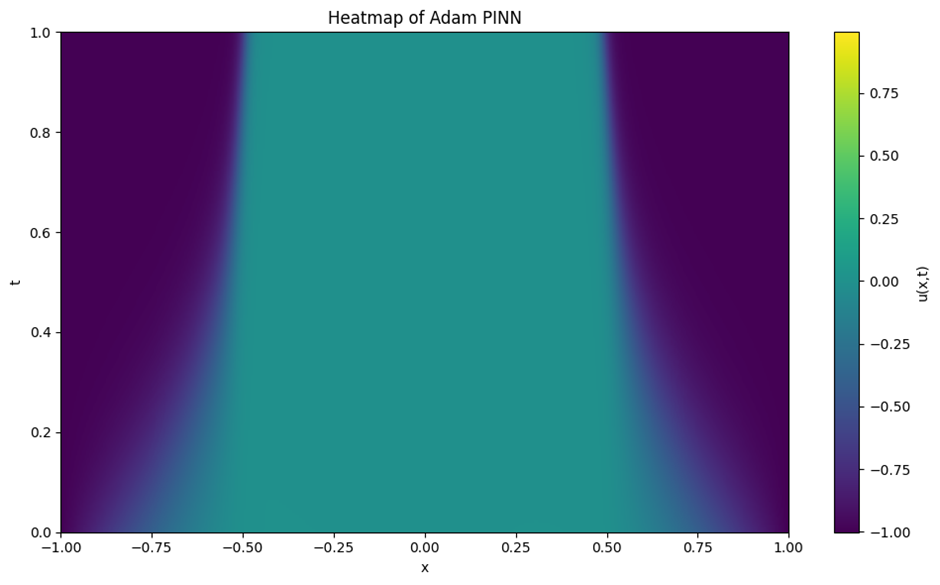}
        \caption{Adam}
    \end{subfigure}
    \hfill
    \begin{subfigure}[t]{0.24\textwidth}
        \centering
        \includegraphics[width=\linewidth]{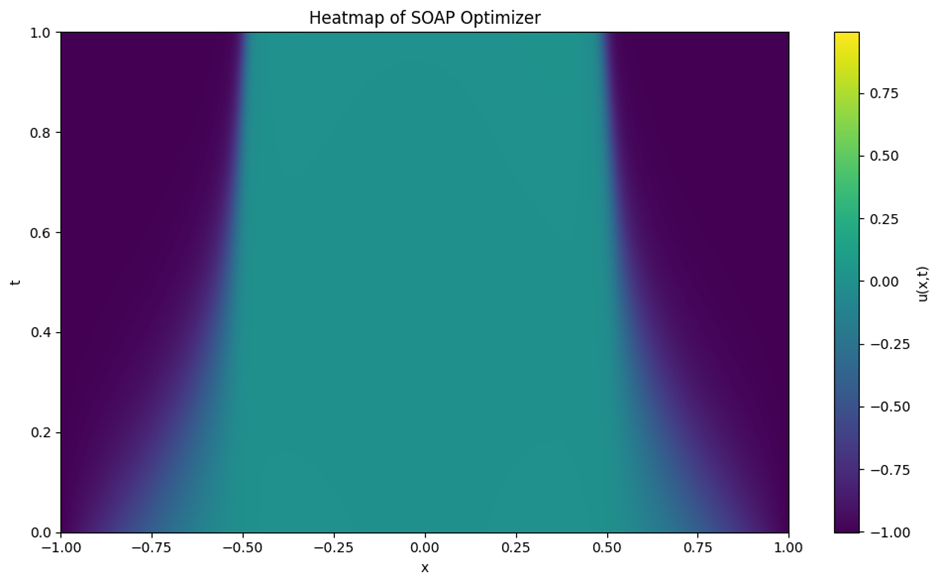}
        \caption{SOAP}
    \end{subfigure}
    \hfill
    \begin{subfigure}[t]{0.24\textwidth}
        \centering
        \includegraphics[width=\linewidth]{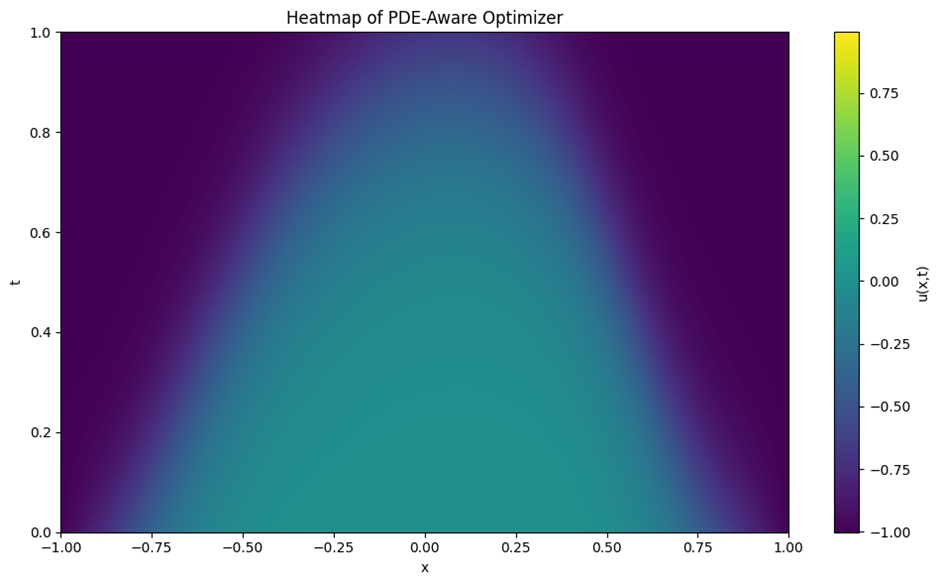}
        \caption{PDE-aware}
    \end{subfigure}
    \caption{Heatmap comparison of Adam, SOAP, and PDE-aware optimizers on the Allen–Cahn equation}
    \label{fig:allen-heatmap-comparison}
\end{figure}

\vspace{1em}

\begin{figure}[H]
    \centering
    \begin{subfigure}[t]{0.32\textwidth}
        \centering
        \includegraphics[width=\linewidth]{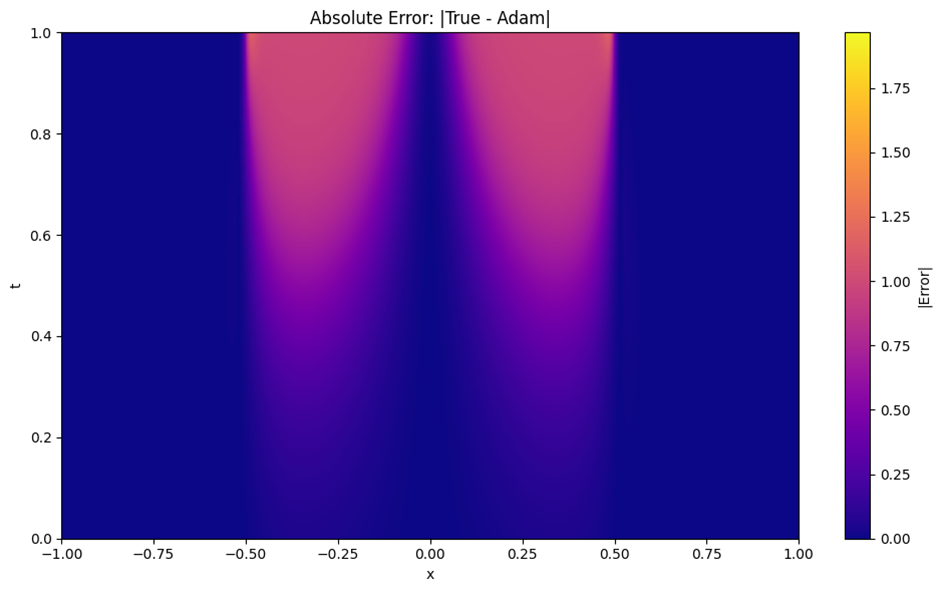}
        \caption{Adam}
    \end{subfigure}
    \hfill
    \begin{subfigure}[t]{0.32\textwidth}
        \centering
        \includegraphics[width=\linewidth]{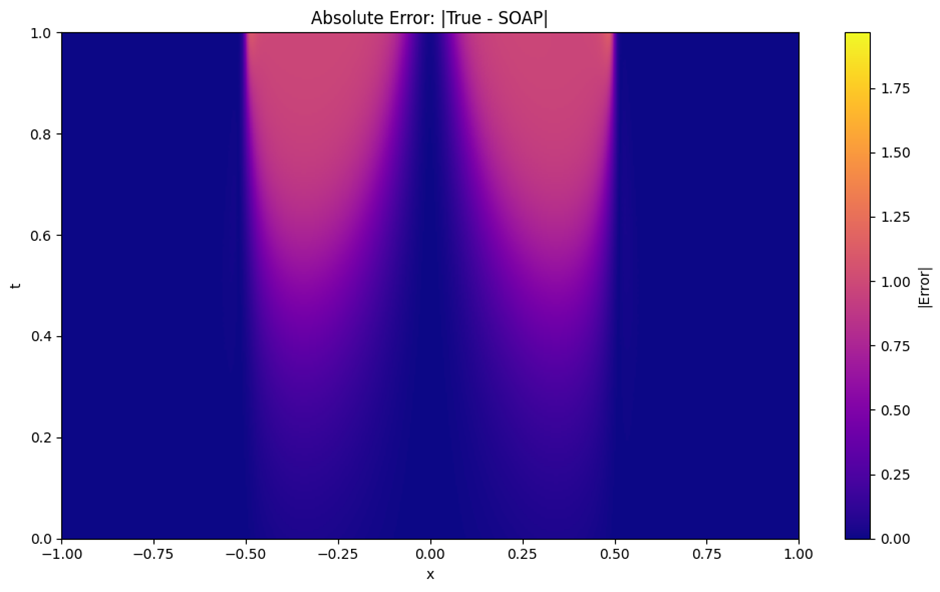}
        \caption{SOAP}
    \end{subfigure}
    \hfill
    \begin{subfigure}[t]{0.32\textwidth}
        \centering
        \includegraphics[width=\linewidth]{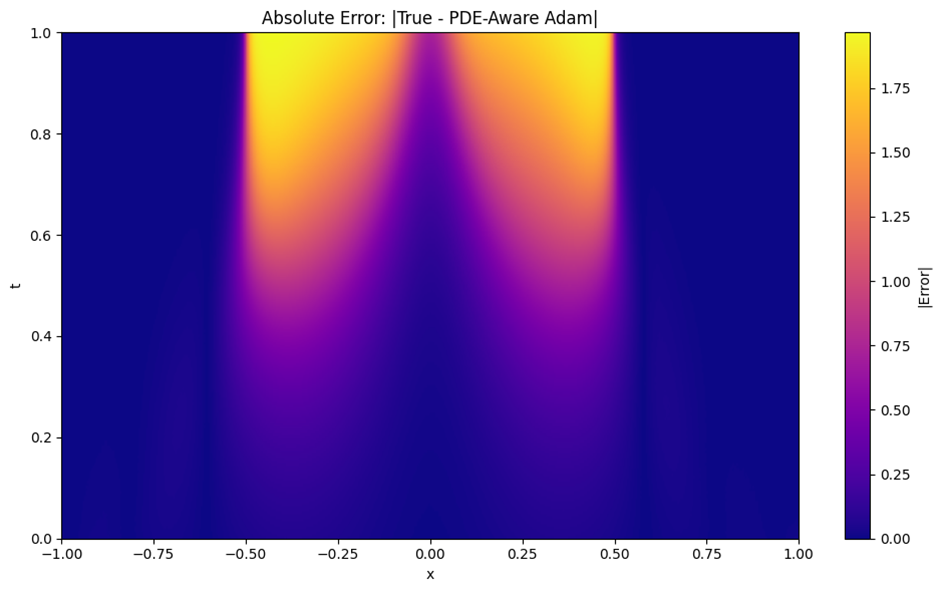}
        \caption{PDE-aware}
    \end{subfigure}
    \caption{Absolute error ($|u_{\text{PINN}} - u_{\text{FDM}}|$) for Allen–Cahn equation across different optimizers}
    \label{fig:allen-absolute-error}
\end{figure}

\vspace{1em}

\begin{figure}[H]
    \centering
    \includegraphics[width=0.4\linewidth]{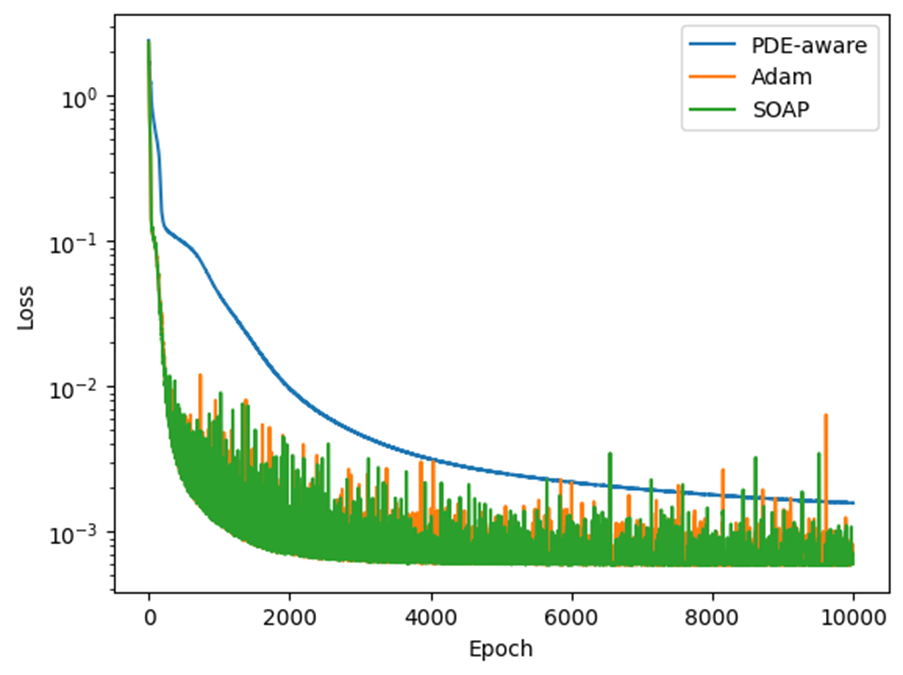}
    \caption{Training loss over epochs for Allen–Cahn equation. PDE-aware optimizer converges smoothly, while Adam and SOAP show faster but noisier behavior.}
    \label{fig:allen-loss}
\end{figure}
\subsection{1D Korteweg–de Vries (KdV) PDE}

\begin{figure}[H]
    \centering
    \begin{subfigure}[t]{0.24\textwidth}
        \centering
        \includegraphics[width=\linewidth]{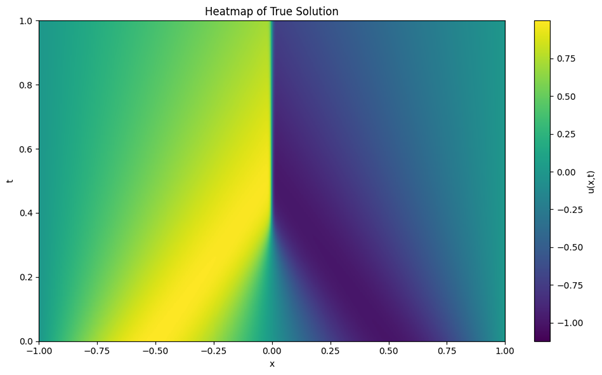}
        \caption{True}
    \end{subfigure}
    \hfill
    \begin{subfigure}[t]{0.24\textwidth}
        \centering
        \includegraphics[width=\linewidth]{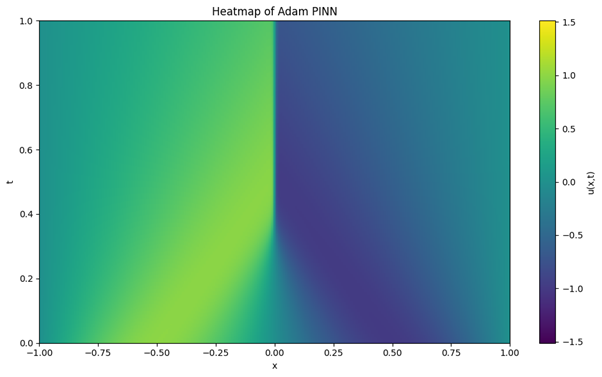}
        \caption{Adam}
    \end{subfigure}
    \hfill
    \begin{subfigure}[t]{0.24\textwidth}
        \centering
        \includegraphics[width=\linewidth]{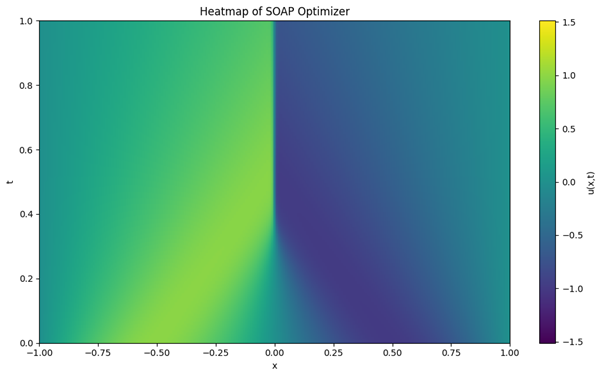}
        \caption{SOAP}
    \end{subfigure}
    \hfill
    \begin{subfigure}[t]{0.24\textwidth}
        \centering
        \includegraphics[width=\linewidth]{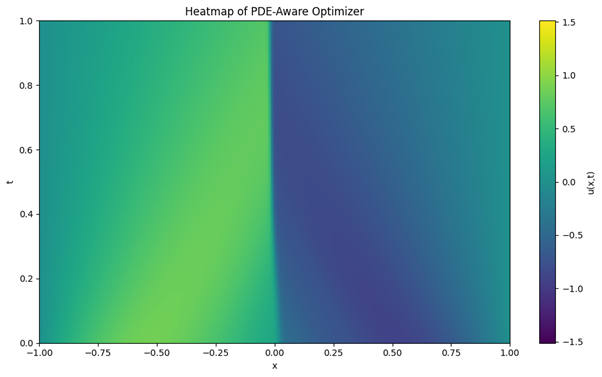}
        \caption{PDE-aware}
    \end{subfigure}
    \caption{Heatmap comparison of Adam, SOAP, and PDE-aware optimizers on the KdV equation}
    \label{fig:kdv-heatmap-comparison}
\end{figure}

\vspace{1em}

\begin{figure}[H]
    \centering
    \begin{subfigure}[t]{0.32\textwidth}
        \centering
        \includegraphics[width=\linewidth]{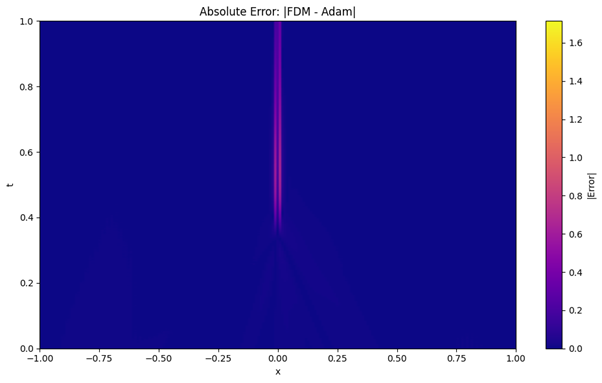}
        \caption{Adam}
    \end{subfigure}
    \hfill
    \begin{subfigure}[t]{0.32\textwidth}
        \centering
        \includegraphics[width=\linewidth]{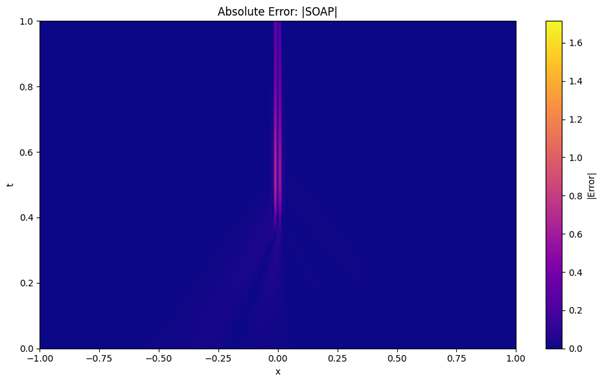}
        \caption{SOAP}
    \end{subfigure}
    \hfill
    \begin{subfigure}[t]{0.32\textwidth}
        \centering
        \includegraphics[width=\linewidth]{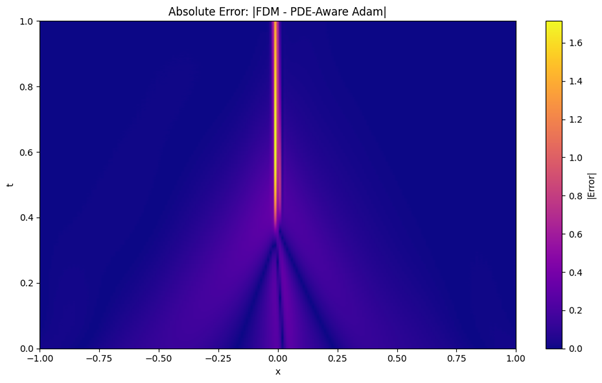}
        \caption{PDE-aware}
    \end{subfigure}
    \caption{Absolute error ($|u_{\text{PINN}} - u_{\text{FDM}}|$) for KdV equation across different optimizers}
    \label{fig:kdv-absolute-error}
\end{figure}

\vspace{1em}

\begin{figure}[H]
    \centering
    \includegraphics[width=0.4\linewidth]{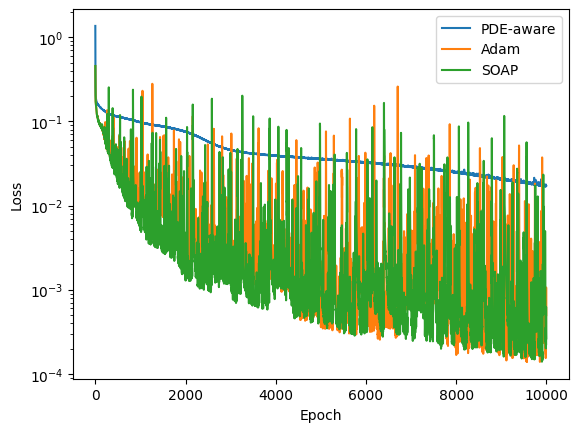}
    \caption{Training loss over epochs for KdV equation. PDE-aware optimizer shows smoother convergence, while Adam and SOAP demonstrate faster but more oscillatory descent.}
    \label{fig:kdv-loss}
\end{figure}

\section{Discussion}

The comparative analysis of Adam, SOAP, and our proposed PDE-aware optimizer across the three benchmark PDEs Burgers, Allen-Cahn, and KdV highlights distinct trade-offs in optimizer behavior. Adam and SOAP demonstrate faster convergence in terms of training loss but exhibit pronounced oscillations during training. In contrast, the PDE-aware optimizer converges more gradually, yet consistently, with significantly smoother loss curves. Qualitatively, the PDE-aware optimizer produces solutions that closely track the true PDE behavior across both benchmarks, particularly in stiff or sharply varying regions. This is further supported by the absolute error plots, where the PDE-aware optimizer yields lower and more uniformly distributed errors compared to its counterparts. While SOAP shows competitive accuracy, it often suffers from instability near sharp gradients.

Overall, the PDE-aware optimizer offers a robust and interpretable alternative by directly adapting to residual gradient variance, which helps mitigate gradient misalignment and improves physical consistency in PINN training. Its performance is especially valuable when modeling stiff systems or when stability and smooth convergence are prioritized.
\section{Limitations}
Our current experiments were conducted using relatively small MLP architectures and limited GPU availability. While results on benchmark PDEs are promising, further validation on larger models and more complex multidimensional systems is needed. Future work should explore scalability and generalization performance using better GPUs and larger models.
\section{Code Availability}
All code and experiments used in this study are publicly available at:\\
\href{https://github.com/vismaychuriwala/PDE-aware-optimzer-jax}{\texttt{https://github.com/vismaychuriwala/PDE-aware-optimzer-jax}}.

\printbibliography
\end{document}